%% file: main_final_submission_TUFFC1.tex
\documentclass[10pt,journal]{IEEEtran}

\input{packages}
\input{macros}

\begin{document}

\title{Pushing the Limit of Unsupervised Learning for Ultrasound Image Artifact Removal}
\author{
Shujaat Khan,
        Jaeyoung Huh,~
        and~Jong~Chul~Ye,~\IEEEmembership{Fellow,~IEEE}
\thanks{The authors are with the Department of Bio and Brain Engineering, Korea Advanced Institute of Science and Technology (KAIST), 
		Daejeon 34141, Republic of Korea (e-mail:\{shujaat,woori93,jong.ye\}@kaist.ac.kr). 
		This work was supported by the National Research Foundation of Korea under Grant NRF-2020R1A2B5B03001980.
		}}

\maketitle

\begin{abstract}
 Ultrasound (US) imaging is a fast and non-invasive  imaging modality which is widely used for real-time clinical imaging applications
 without concerning
 about radiation hazard.
Unfortunately, it often suffers from poor visual quality from various origins, such as speckle noises, blurring,  multi-line acquisition (MLA),
limited RF channels, small number of view angles for the case of plane wave imaging, etc.
Classical methods to deal with these problems include image-domain signal processing approaches using
various adaptive filtering and model-based approaches. 
Recently, deep learning approaches have been successfully used for ultrasound imaging field.
However, one of the limitations of these approaches is that paired high quality images for supervised training are
 difficult to obtain in many practical applications.
In this paper,  inspired by the recent theory of unsupervised learning using
optimal transport driven cycleGAN (OT-cycleGAN),
we investigate
 applicability of unsupervised deep learning for  US artifact removal problems
without matched reference data.
Experimental results for 
various tasks such as deconvolution, speckle removal,  limited data artifact removal,  etc.
confirmed that our unsupervised learning method provides comparable results to supervised learning for many practical applications.
\end{abstract}

\begin{IEEEkeywords}
Ultrasound imaging, Portable, 3D, low-powered, Deep Learning, Sparse Sampling, Inverse Problem, Deconvolution.
\end{IEEEkeywords}

\IEEEpeerreviewmaketitle

\section{Introduction}
\label{sec:introduction}
Ultrasound (US) imaging is a safe imaging modality with high temporal resolution,
so it is considered as a first choice for various clinical applications such as echo-cardiography, fetal scan, etc. 

To form an US  image,  individual channel RF measurements are back-propagated and accumulated after applying specific delays \cite{TimeReversal1}. 
Accordingly, the quality of US images is limited by number of factors such as inhomogeneous sound speed,  limited number of channels,  frame rate,  etc. For instance, in conventional focused B-mode ultrasound, lateral and axis resolutions depend on the number of scan-lines and axial sampling frequency,
whereas in planewave B-mode imaging the the quality of image is defined by the number of planewaves used in coherent-planewave compounding (CPC) \cite{montaldo2009coherent}.   In addition,  sonographic signals are inherently susceptible to speckle noises, which appear as  granule patterns in US images  and  reduce the contrast-to-noise ratio (CNR) significantly \cite{zhang2020despeckle}.

 To address these issue, model-based methods have been developed to remove the noises. These methods include
 classical methods such as 
adaptive filtering, deconvolution,  etc.\cite{NLLR_Despeckle, zhang2020despeckle, coupe2009nonlocal,chen2015compressive, jensen1992deconvolution, 7565583, schretter2017ultrasound}. 
However, these methods are usually associated with high computation cost 
and may be subject to the model inconsistency \cite{chen2015compressive, jensen1992deconvolution, 7565583, schretter2017ultrasound}.  


A promising direction to mitigate these issues is a deep learning approach. In the recent past, deep learning has emerged as a promising tool for variety of medical imaging related inverse  problems \cite{kang2017deep, ye2017deep, han2017deep, aggarwal2017modl, khan2019deep, wurfl2016deep}. In particular, for high-quality ultrasound imaging there are number of solutions proposed by various researchers, which can be categorized into i) channel data based solutions \cite{yoon2018efficient,khan2020adaptive}, and ii) image domain based solutions \cite{feigin2018deep, yoon2018efficient, nair2018fully, khan2019universal, jafari2020cardiac, vedula2018high, nair2019generative, yu2018deep, ozkan2018inverse, cohen2018sparse}.
 Typically, channel domain approaches
such as  deep beamformers \cite{khan2020adaptive} show better generalization in diverse imaging conditions, but
 image domain approaches are more easier to implement 
for various applications without accessing the channel data.

%
%
%
%
%

The 
existing deep learning strategies mostly rely on the paired dataset for supervised training. However, in many real world imaging situations, access to paired images (input image and its desired output pair) is not possible. For example, to improve the visual quality of US images acquired using a low-cost imaging system we need to scan exactly the same part using a high-end machine, which is not possible.  
In another example,
the conversion of the plane wave imaging to the focused B-mode image quality is not applicable by supervised learning
as we should obtain the same image using two different
acquisition modes.
One could use simulation data for supervised training, but it is prone to bias to simulation environment.
Therefore, we are interested in developing an unsupervised learning strategy where low quality images from one scan can be used as inputs, whereas high quality images from different anatomy and imaging conditions can be used as target images, by doing so we can successfully train an image enhancement model. 

We are aware that there are recent approaches in US literature that aim at similar unsupervised learning set-up \cite{jafari2020cardiac, huang2019mimicknet}. For example, the authors in \cite{jafari2020cardiac} employed cycleGAN network to improve the image quality from portable US image using high-end unmatched image data to improve the accuracy of cardiac chamber segmentation. Similar image domain cycleGAN approach has been recently proposed \cite{huang2019mimicknet}.
However, it is not clear whether such quality improvement is real or a cosmetic change.
Moreover, although there are potentially many applications beyond the conversion of the low-end image to high-end image,
these additional applications have been never investigated.

Therefore, one of the most important contributions of this study is an application
of recent theory of unsupervised learning using the optimal transport driven cycleGAN (OT-cycleGAN)  \cite{Reference:sim2019optimal} for unsupervised artifact removal. Unlike the black-box application of cycleGAN, the OT-cycleGAN
 was derived using optimal transport theory \cite{villani2008optimal,peyre2019computational} that transports a probability distribution of noisy images
to clean image distribution  \cite{Reference:sim2019optimal}. Therefore, if properly trained, the theory guarantees that
the image  improvement is not a cosmetic changes, but a real improvement by learning the distribution of the clean image data distributions.
Another important contribution of this paper is the extension of unsupervised learning to 
various US artifact removal problems, such as speckle noise removal, deconvolution, limited measurements
artifact removal,  plane wave to focused B-mode imaging conversion, etc, which verify that
our method provides near comparable results  in both qualitative
and quantitative manners. Furthermore, our framework is so general that it can be used for various applications of
US image quality improvement without concerning about collecting paired reference data.

The rest of the paper is organized as follows. In Section \ref{sec:theory},  we briefly
review the recent theory of OT-cycleGAN for unsupervised deep learning applications.
The 
detailed information of the proposed method are presented in Section \ref{sec:methods}, which
is  followed by the results and discussion in Section \ref{sec:results}. Finally the paper is concluded in Section \ref{sec:conclusion}.

\section{Theory}
\label{sec:theory}

Here, we briefly introduced OT-CycleGAN \cite{Reference:sim2019optimal} to make the paper self-contained.
However, our derivation is different from \cite{Reference:sim2019optimal}, since here we focus on the general geometry
of unsupervised learning and  explain why cycleGAN is a natural way to address this problem.

\subsection{Wasserstein Metric and Optimal Transport}

{Optimal transport (OT) provides a mathematical means to compare two probability measures} \cite{villani2008optimal,peyre2019computational}. 
 Formally, we say that  $T:\Xc \mapsto \Yc$ transports the probability measure $\mu \in P(\Xc)$ to another measure $\nu \in  P(\Yc)$, if
\begin{eqnarray}\label{eq:constraint}
\nu(B) = \mu\left(T^{-1}(B)\right),\quad \mbox{for all $\nu$-measurable sets $B$},
\end{eqnarray}
{Suppose there is a cost function $c:\Xc \times \Yc \rightarrow \Rd\cup\{\infty\}$ such that $c(x,y)$ represents the cost of moving one unit of mass from $x \in \Xc$ to $y \in \Yc$.}
Monge's original OT problem \cite{villani2008optimal,peyre2019computational} is then to find a transport map $T$ that transports $\mu$ to $\nu$
at the minimum total transportation cost.
Kantorovich  relaxed the assumption to consider probabilistic
transport that allows  mass splitting from a source
toward several targets:
\begin{align}\label{eq:OTcost}
&\min_{\pi \in \Pi (\mu, \nu)} \int_{\Xc \times \Yc} c(x,y) d\pi(x,y)
\end{align}
where  $\Pi(\mu, \nu)$ is the set of joint distributions whose marginal distribution is $\mu$ and $\nu$, respectively.
%


If we choose a metric $d$  in $\Xc$ as a transportation cost $c$, then 
the optimal transport cost in \eqref{eq:OTcost} becomes
 Wasserstein-1 distance between two probability measures $\mu$  and $\nu$:
\begin{align}\label{eq:Wp}
W_1(\mu,\nu):= &\inf\limits_{\pi \in \Pi(\mu,\nu)}\int_{\Xc\times \Yc} d(x,y)d\pi(x,y) \\
=& \inf\limits_{\pi \in \Pi(\mu,\nu)} \Ed_\pi\left[d(X,Y)\right]
\end{align}
where 
$X,Y$ are the random vectors with  the joint distribution $\pi$,
and $\Ed_\pi[\cdot,\cdot]$ is the expectation with respect to the joint measure $\pi$.
Therefore, the meaning of the Wasserstein-1 metric is that
the minimum average distance between samples in two probability distributions $\mu$ and $\nu$,
the optimal transport theory is concerned about minimizing the average distance.

Unlike Kullback–Leibler (KL) divergence \cite{kullback1997information},
Wasserstein metric is a real metric that satisfies all  properties of  a metric in the metric space \cite{villani2008optimal,peyre2019computational}. Therefore, it provides a powerful way of measuring
distance in the probability space, which is useful for unsupervised learning as described in the next section.

\subsection{Geometry of Unsupervised Learning}

\begin{figure}[!hbt] 	
\center{ 
\includegraphics[width=5cm]{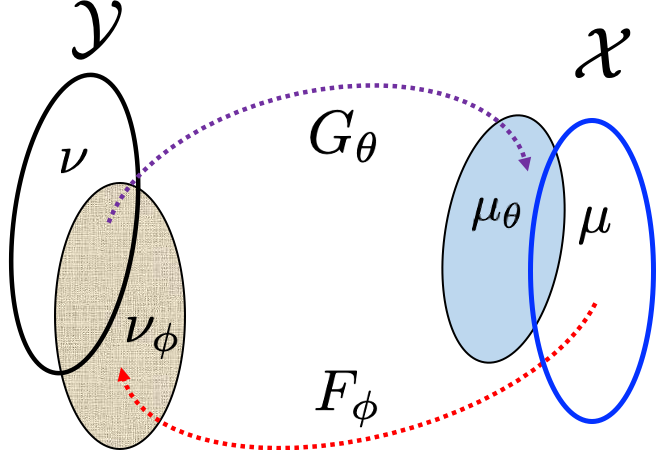}
}
\caption{Geometric view of unsupervised learning.}
\label{fig:cycleGANgeom}
\end{figure}

Our geometric
view of unsupervised learning is shown in Fig.~\ref{fig:cycleGANgeom}.
Here, the target image space
$\Xc$ is equipped with a probability measure $\mu$, whereas
the original image space is $\Yc$ with a probability measure $\nu$.
Since there are no paired data, the goal of unsupervised learning is to match the probability distributions rather than each individual samples.
This can be done by finding  transportation maps that transport the measure $\mu$ to $\nu$, and vice versa.

More specifically, the transportation from a measure space $(\Yc,\nu)$ to another measure space $(\Xc,\mu)$ is done by a generator $G_\theta: \Yc \mapsto \Xc$, realized by a deep
network parameterized with $\theta$.  Then, the generator $G_\theta$ ``pushes forward'' the measure $\nu$ in $\Yc$ to a meaure $\mu_\theta$ in the target space $\Xc$ \cite{villani2008optimal,peyre2019computational}. Similarly, the transport from $(\Xc,\mu)$ to $(\Yc,\nu)$ is performed by another neural network generator $F_\phi$,
 so that the generator $F_\phi$ pushes forward the measure $\mu$ in $\Xc$ to $\nu_\phi$ in the original space $\Yc$.
Then, the optimal transport map for unsupervised learning can be achieved by minimizing the statistical distances between $\mu$ and $\mu_\theta$, and
 between $\nu$ and $\nu_\phi$, and our proposal is to use the Wasserstein-1 metric as a means to measure the statistical distance.

More specifically, for the choice of a metric $d(x,x')=\|x-x'\|$ in $\Xc$,  using the change of measure formula \cite{villani2008optimal,peyre2019computational},
the Wasserstein-1 metric between $\mu$ and $\mu_\theta$ can
be computed by
\begin{align}\label{eq:Wmu}
W_1(\mu,\mu_\theta)
=&\inf\limits_{\pi \in \Pi(\mu,\nu)}\int_{\Xc\times \Yc} \|x-G_\theta(y)\|d\pi(x,y) 
\end{align}
Similarly, the Wasserstein-1 distance between $\nu$ and $\nu_\phi$ is given by
\begin{align}\label{eq:Wnu}
W_1(\nu,\nu_\phi)
=&\inf\limits_{\pi \in \Pi(\mu,\nu)}\int_{\Xc\times \Yc} \|F_\phi(x)-y\|d\pi(x,y) 
\end{align}

Here, care should be taken,  since we should minimize the two statistical distances simultaneously for unsupervised learning.
More specifically, rather than minimizing \eqref{eq:Wmu} and \eqref{eq:Wnu} separately with distinct joint distributions,
a better way of finding the transportation map is to minimize them together with the same joint distribution $\pi$:
\begin{align}\label{eq:unsupervised}
\inf\limits_{\pi \in \Pi(\mu,\nu)}\int_{\Xc\times \Yc} \|x-G_\theta(y)\|+ \|F_\phi(x)-y\|d\pi(x,y) 
\end{align}
This is our unsupervised learning formulation from optimal transport perspective \cite{Reference:sim2019optimal}.

\subsection{Optimal transport driven cycleGAN (OT-CycleGAN)}

One of the most important contributions of our companion paper \cite{Reference:sim2019optimal} is to show that
the primal formulation of the unsupervised learning in \eqref{eq:unsupervised} 
can be represented by a dual formulation. More specifically, the following primal problem
\begin{align}\label{eq:primal}
\min_{\theta,\phi}\inf\limits_{\pi \in \Pi(\mu,\nu)}\int_{\Xc\times \Yc} \|x-G_\theta(y)\|+ \|F_\phi(x)-y\|d\pi(x,y) 
\end{align}
is equivalent to the following dual formulation which we call the optimal
transport driven CycleGAN (OT-cycleGAN):
\begin{eqnarray}\label{eq:OTcycleGAN}
\min_{\phi,\theta}\max_{\psi,\varphi}\ell_{cycleGAN}(\theta,\phi;\psi,\varphi)
\end{eqnarray}
where 
\begin{eqnarray}
\ell_{cycleGAN}(\theta,\phi):=  \gamma \ell_{cycle}(\theta,\phi) +\ell_{Disc}(\theta,\phi;\psi,\varphi) 
\end{eqnarray}
where $\gamma>0$ is the hyper-parameter, and  the cycle-consistency term is given by
\begin{align*}
\ell_{cycle}(\theta,\phi)  =& \int_{\Xc} \|x- G_\theta(F_\phi(x)) \|  d\mu(x) \\
&+\int_{\Yc} \|y-F_\phi(G_\theta(y))\|   d\nu(y)
\end{align*}
whereas  the second term is
\begin{align*}
&\ell_{Disc}(\theta,\phi;\psi,\varphi)  \\
=&\max_{\varphi}\int_\Xc \varphi(x)  d\mu(x) - \int_\Yc \varphi(G_\theta(y))d\nu(y)  \\
 & + \max_{\psi}\int_{\Yc} \psi(y)  d\nu(y) - \int_\Xc \psi(F_\phi(x))  d\mu(x) \notag
\end{align*}

 Here, $\varphi,\psi$ are often called Kantorovich potentials and satisfy 1-Lipschitz condition (i.e.
\begin{align*}
|\varphi(x)-\varphi(x')|\leq \|x-x'\|,&~\forall x,x'\in \Xc \\
|\psi(y)-\psi(y')|\leq \|y-y'\|,&~\forall y,y'\in \Yc
\end{align*}
In machine learning context,  the 1-Lipschitz potentials  $\varphi$ and $\psi$
correspond to the  Wasserstein-GAN (W-GAN) discriminators \cite{arjovsky2017wasserstein}.
Specifically, 
$\varphi$ tries to find the difference between the true image $x$ and the generated image $G_\Theta(y)$,
whereas $\psi$ attempts to find the fake measurement data  that are generated by the synthetic
measurement procedure $F_\phi(x)$.
In fact, this formulation is equivalent to the cycleGAN formulation \cite{zhu2017unpaired} except for the use of 1-Lipschitz discriminators.

Here, care should be taken  to ensure that the Kantorovich potentials  become 1-Lipschitz.
There are many approaches to address this. For example, in the original W-GAN paper \cite{arjovsky2017wasserstein}, the weight
clipping was used to impose 1-Lipschitz condition.
Another method is to use the spectral normalization method  \cite{miyato2018spectral}, which utilizes the power iteration method to impose
constraint on the largest singular value of weight matrix in each layer.
Yet another popular method is the  WGAN with the gradient penalty (WGAN-GP), where
the gradient of the Kantorovich potential is constrained to be 1 \cite{gulrajani2017improved}.
Finally, in our companion paper \cite{lim2019cyclegan}, we also showed that the popular LS-GAN approach \cite{mao2017least}, which is often used in combination of standard
cycleGAN \cite{zhu2017unpaired},  is also closely related to
imposing the 1-Lipschitz condition.
In this paper, we  therefore consider LS-GAN variation as our implementation for discriminator term where
the discriminator loss is given by
\begin{align}\label{eq:ourLSnonblind}
&\ell_{Disc}(\theta,\phi;\psi,\varphi)=  \notag \\
&-\int_\Xc  (\varphi(x)-1)^2d\mu(x) - \int_\Yc \left(\varphi(G_\Theta(y))+1\right)^2d\nu(y) \notag\\
&-\int_\Yc  (\psi(y)-1)^2d\nu(y) - \int_\Xc \left(\varphi(F_\theta(x))+1\right)^2d\mu(x) \notag\\
\end{align}

\subsection{Unsupervised US  Artifact Removal}

Based on the mathematical background of unsupervised learning
and its implementation using OT-cycleGAN,
we are interested in solving the following unsupervised learning problems in US:
\begin{enumerate}
\item Deconvolution.
\item Speckle noise removal.
\item Planewave image enhancement using high quality focused B-mode target.
\item Multi-line acquisition (MLA) block artifact and noise removal in cardiac imaging.
\item Missing channel artifact removal
\end{enumerate}
For each type of enhancement, we generated target data starting from the fully-sampled RX data i.e., $64$ channels from unmatched data set. The details of each application is given below.

\subsubsection{Deconvolution Ultrasound}
The axial resolution of ultrasound imaging is limited by the bandwidth of the transducer.  Conventional beamforming methods such as delay-and-sum (DAS)  are limited by the accuracy of ray approximation of the wave propagation \cite{jensen1992deconvolution}. In order to overcome these issues, many researchers have explored the deconvolution of US images \cite{7565583, chen2015compressive,jensen1992deconvolution}. Deconvolution ultrasound may help in dealing with modeling inaccuracies and finite bandwidth issues, which will eventually improve the spatial resolution of an ultrasonic imaging system.  

Specifically, the received signal is modeled as a convolution of tissue reflectivity function (TRF) $x$ with a point spread function (PSF) $h$, where tissue reflectivity function represents scatter's acoustic properties, while the impulse response of the imaging system is modeled by point spread function. 
The estimation of $x$ from the DAS measurement  is known as a deconvolution US problem. In most practical cases, the complete knowledge of $h$ is not available, and therefore both unknown TRF $x$ and  the PSF $h$ have to be estimated together, which is called the blind deconvolution problem \cite{taxt1999noise}. One strategy is to estimate $h$ and $x$ jointly \cite{yu2012blind}, and another strategy is to estimate them separately \cite{jensen1991estimation}, i.e., first $h$ is estimated from $y$, and then $x$ is estimated based on $h$ \cite{jirik2008two}. 

In this paper, for unmatched target distribution data generation,  the second strategy is used first with small number of DAS images, which is followed by training the  DeepBF \cite{khan2020adaptive, 7565583} to perform deconvolution-based beamforming.  This way a large number of target data can be
generated easily without solving deconvolution problems for large number of data set.
%

\subsubsection{Speckle-noise removal}

The granular patterns appears in US images due to constructive and destructive interference of ultrasonic wave. These are called `speckle' noise. The speckle noise is a multiplicative impulse noise. It is a major reason of quality degradation and removal of it can improve the visual quality and subsequently enhance the structural details in US images \cite{NLLR_Despeckle}. In recent past, a variety of reasonably good de-speckling algorithms have been proposed for US imaging \cite{NLLR_Despeckle, coupe2009nonlocal, zhang2020despeckle}. However, most of them are either too slow to use for run-time application or require complicated configurations of parameter for each image. These issues hinder the utilization in real world scenarios. 

 One such algorithm is proposed by Zhu \textit{et al} \cite{NLLR_Despeckle}, which is based on the principal of non-local low-rank (NLLR) filtering. To generate speckle free target data herein, we used NLLR method. In NLLR, the image is pre-processed to generate a guidance map and later non-local filtering operations are performed on the candidate patches that are selected using that guidance map.  For further refinement of filtered patches, a truncated nuclear norm (TWNN) and structured sparsity criterion are used \cite{zhang2012matrix, gu2014weighted}. This algorithm
 is used to generate our target samples for despeckle images.


\subsubsection{Planewave image enhancement using high quality focused B-mode target}

Planewave (PW) imaging is an emerging mode of US scanning. It offers ultra-fast scanning capabilities with comparable image quality. In PW imaging the quality of the acquired scan depends on the number of planewaves (PWs) used to generate the final image.
For most of the clinical applications, multiple PWs are combined using the CPC method to produce a desired quality image. However, there is a trade-off in the quality and speed of the scan as each PW scanning require additional scanning time,  limiting the application of PWI for high quality accelerated imaging \cite{tanter2014ultrafast}. 

To find an optimal trade-off between speed and visual quality, there are number of deep learning based PW compounding methods  \cite{gasse2017high, khan2019universal}. However, these method require access to fully-sampled ($31\sim75$) planewaves data to train a supervised model. Typical lower-end commercial systems are not equipped with such hardware complexity to produce such a high quality label dataset. 
 
 Therefore, we propose to use an unsupervised learning in which high-quality label images are obtained using focused B-mode imaging. 
For further quality improvement, the target focused B-mode images are processed using deconvolution and filtering with NLLR \cite{NLLR_Despeckle} speckle denoising algorithm.


\subsubsection{MLA block artifact and noise removal in cardiac imaging}
Echocardiography (ECHO) require fast scan time, and it is typically performed by a phased array probe operating in focused scanning mode in which multiple scan-lines are combined to form a complete image. Therefore, to scan a large region of interest, high number of scan-lines are required resulting in reduced temporal resolution. 

For accelerated echocardiography, conventional acceleration methods like multi-line acquisition (MLA) are used, where each transmit/receive event's data is used to generate multiple scan-lines. The limitation of the MLA is that it works only for limited acceleration factor and produces blocking artifacts for high frame rate \cite{matrone2017high,vedula2018high}.  
In addition to measurement's limitation, sonographic signals from echocardiography are susceptible of speckle noise which is also a major factor for the degradation of visual quality.
 As such, this visual quality and temporal resolution trade-off is a bottle neck for many echocardiography applications.


A variety of deep learning based block artifact removal methods exist, but they are designed for supervised training and require access to high quality labelled channel data \cite{yoon2018efficient,matrone2017high,vedula2018high}. In this study we proposed an image domain unsupervised MLA artefact and noise removal deep neural network method. Unpaired target image distributions are generated with single-line-acquisition (SLA) and filtered using NLLR \cite{NLLR_Despeckle} filtering technique.

\subsubsection{ Missing channel artifact removal}
The power consumption, size, and cost of the US system are mainly dependent on the number of measurement channels. Therefore, in portable and three dimension ultrasound imaging system, there is an increasing demand for  computational algorithms which can produce high quality images using fewer receive channels. Conventional beamforming methods are not designed for sub-sampled RF data and standard DAS is highly susceptible to sub-sampling in measurements. On the other hand, advance compressive beamforming methods \cite{wagner2012compressed,burshtein2016sub,cohen2018sparse} are computationally expensive and require hardware modifications limiting their use as generalized solution.  

Recently a deep learning based compressive beamformer was proposed \cite{khan2020adaptive}, which can help reconstruct high quality images from limited measurements. However, the method in \cite{khan2020adaptive} requires an access to fully-sampled label data which is not accessible for low-cost imaging system. In this study, we proposed to design an image domain quality enhancement method that can directly process corrupted images to remove missing channel artefact, and improve the contrast and resolution of the B-mode images. 

For the generation of unpaired target data distribution,
unmatched  target images are generated using DeepBF \cite{khan2020adaptive}. 


\begin{figure*}[!hbt]
	\centerline{\includegraphics*[width=0.8\textwidth]{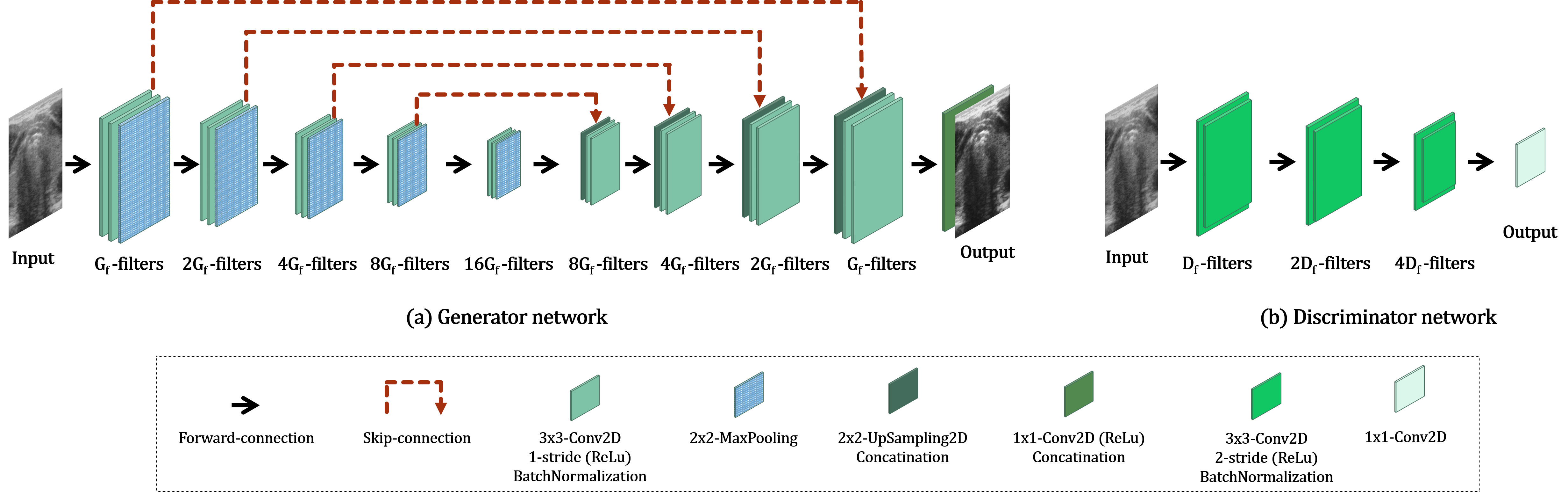}}
	\caption{Proposed network architecture: (a) Generator network, (b) Discriminator network.}
	\label{fig:architecture}	
\end{figure*}

\section{Method}
\label{sec:methods}

\subsection{Dataset}
In this study, we used four different dataset,  all were acquired using an E-CUBE 12R US system (Alpinion Co., Korea).  For data acquisition, we used a linear array (L3-12H), and phased array (SP1-5) transducers and their configuration are given in Table \ref{probe_config}. 
\begin{table}[!hbt]
	\centering
	\caption{US Probes Configuration}
	\label{probe_config}
	\resizebox{0.48\textwidth}{!}{
		\begin{tabular}{c|c|c}
			\hline
			{Parameter} & {Linear array} & {Phased array} \\ \hline\hline
			Probe Model No. & L3-12H &  SP1-5 \\
			Carrier wave frequency & 8.48/10.0 MHz & 3.1 MHz\\
			Sampling frequency & 40 MHz & 40 MHz\\
			Scan wave mode & Focused/ 31-Planewaves & Focused\\
			No. of probe elements & 192 & 192\\
			No. of Tx elements & 128 & 128\\
			No. of TE events & 96 & 96 \\
			No. of Rx elements & 64 (from center of Tx) & 64 (from center of Tx)\\
			Elements pitch & 0.2 mm & 0.3 mm\\
			Elements width & 0.14 mm & 0.22 mm\\
			Elevating length & 4.5 mm & 13.5 mm\\
			Axial depth range & 20$\sim$80 mm & 75 mm\\
			Lateral length & 38.4 mm & 57.6 mm \\
			Focal depth range & 10$\sim$40 mm & 45 mm\\\hline
	\end{tabular}}
\end{table}

\subsubsection{Linear array focused B-mode dataset}
The first data consist of $400$ \textit{in-vivo} and $218$ phantom frames scanned using a center frequency of $8.48$ MHz. The \textit{in-vivo} dataset acquired from the carotid/thyroid area of $10$ volunteers, $40$ temporal frames were scanned from each subject. For phantom dataset, we acquired $218$ frames from ATS-539 multipurpose tissue mimicking phantom. The phantom was scanned from different views angles. Second dataset was scanned from the calf and forearm regions of two volunteers using $10$ MHz carrier frequency. There are total $100$ images were scanned $50$ from each body part.

For deconvolution and denoising experiments, the training was performed using only the first data. In particular, for training purpose the dataset of $8$ individuals consist of $320$ in-vivo and $192$ phantom images were used, while remaining $80$ images from two different individuals and $26$ images from completely different region of phantom were used for testing. For additional validation, second independent dataset is used. Note that all models were trained on same $508$ images and no additional training is performed on any of the testing or independent dataset.

For missing channel artifact experiments the above mentioned data set is expanded into six subsets of input data each consist of $4$, $8$, $16$, $24$, $32$, and $64$ channels representing $16\times$, $8\times$, $4\times$, $2.667\times$, $2\times$, and $1\times$ sub-sampling rates respectively. 

\subsubsection{Linear array planewave B-mode dataset}
For planewave imaging experiments, we collected the third dataset. For this dataset, we used the same (L3-12) operating at center frequency of $8.48$ MHz in a planewave mode. There are total $309$ scans acquired, $100$ from ATS-539 phantom and $209$ from \textit{in-vivo} carotid/thyroid area of $10$ volunteers. The dataset is expanded by using different subsets of PWs to simulate different imaging configurations. In particular, four subsets were used each consist of $31$PWs, $11$PWs, $7$PWs and $3$PWs. For planewave image enhancement experiments $508=127\times4$ images were used. All PW images were processed using standard DAS and CPC method \cite{montaldo2009coherent}. The training dataset is composed of $127$ images $50$ of which were acquired from ATS-539 phantom and remaining $77$ from \textit{in-vivo} dataset of $4$ volunteers.  The remaining dataset of phantom and $6$ volunteers was used for testing purpose only.

\subsubsection{Phased array B-mode dataset}
To design MLA artifact removal experiment, we designed an additional dataset. This dataset was acquired using (SP1-5) phased array probe and it consist of $105$ scans of different regions of ATS-539 phantom and $489$ scans from the cardiac region of $7$ volunteers. Five subsets of images are generated using $16$, $24$, $32$, $48$ and $96$ transmit events representing $6$-MLA, $4$-MLA, $3$-MLA, $2$-MLA and a single line acquisition (SLA). The target images are generated from SLA images filtered with NLLR \cite{NLLR_Despeckle} algorithm. The training dataset used in this study consists of $55$ phantom scans, and $297$ \textit{in-vivo} scans acquired from $5$ individual, while remaining dataset were used for testing. Total number of images in training and test datasets are $1760=5\times(55+297)$, and $1210=5\times(50+192)$ respectively.

\subsection{Network specification}


\subsubsection{Generator Model}
The generator model has a U-Net architecture as shown in Fig.~\ref{fig:architecture}(a). The model comprises of $9$ modules, which consists of $27$ convolution layers with batch-normalization, pooling, up-sampling and concatenation blocks for skip connections.  For all convolution layers ReLu activation function and 2D filters of kernel size $(3\times3)$ were used, except for the output layer, where $(1\times 1)$ filter size is used. The number of filters $G_f$ is doubled in every next module of encoder part and halve in every next decoder module, expect for the output layer where only a single layer is used to produce single channel output. For example, in deconvolution US and despeckle (speckle noise removal) experiments the number of channels starts from $8$ i.e., in the first module there were $8$ channels (number of filters) and in the next module the number of filters increased to $16$, $32$ and so forth. For missing channel and MLA artifact removal experiments, the number of channels starts from  $64$ and $16$ respectively. 

\subsubsection{Discriminator Model}
The discriminator model is a fully convolution neural network model to implement PatchGAN \cite{zhu2017unpaired}. Unlike conventional discriminator where mapping between an input image to a single scalar vector is performed, PatchGAN learns the mapping of sub-array (Patch) representing individual elements and their relative position in an image. The model comprises of $4$ convolution blocks each consist of a set of two convolution layers having stride-size of $2$ with batch normalization and Leaky ReLu activation function. The number of filters $D_f$ are doubled in every next module, expect for the output layer where only a single layer is used to produce single channel output. The filter size in all layers was again $(3\times3)$ except for the last layer where $(1\times1)$ filter size is used. For all experiments the number of filter was $256$,  except for the MLA artifact experiment where it was chosen to be $512$. A detailed schematic of discriminator model is shown in Fig.~\ref{fig:architecture}(b).

%

\subsection{Performance metrics}

%

For quantitative evaluation of our proposed method, the standard quality metrics of ultrasound imaging are used.
Specifically, as the   local anatomical structure or region of interest are important in US, 
 we used contrast statistics. The contrast between the two regions of interests ($R_a$) and ($R_b$) in the image is quantified in terms of contrast-recovery (CR), contrast-to-noise ratio (CNR), and generalized CNR (GCNR) \cite{rodriguez2019generalized}. To select region ($R_a$) and ($R_b$), we manually generated separate ROI masks for each image. 
 
More specifically, the contrast recovery is quantified as
\begin{equation}
{\hbox{CR}}(R_a,R_b) = |\mu_{R_a}-\mu_{R_b}|
\end{equation}
where $\mu_{R_a}$, and $\mu_{R_b}$, are the local means of region ($R_a$) and ($R_b$) respectively.
The CR measure is a standard measure for contrast. However, it does not consider the SNR loss. In typical contrast enhancement methods, the contrast is usually improved at the cost of SNR; therefore, to estimate the overall gain in contrast with respect to noise level, we used CNR measure which is defined as
\begin{equation}
{\hbox{CNR}}(R_a,R_b) = \frac{|\mu_{R_a}-\mu_{R_b}|}{\sqrt{\sigma^2_{R_a} + \sigma^2_{R_b}}},
\end{equation}
where $\sigma_{R_a}$, and $\sigma_{R_b}$ are the standard deviations of region ($R_a$) and ($R_b$) respectively.
Recently, a more reliable measure of contrast is proposed called generalized-CNR (GCNR) \cite{rodriguez2019generalized}. The GCNR is supposed to be an unbiased measure of contrast in which the overlap between the intensity distributions of two regions are compared as
\begin{equation}
{\hbox{GCNR}}(R_a,R_b) = 1- \int \min \{p_{R_a} (i), p_{R_b} (i) \} di,
\end{equation}
where $i$ is the pixel intensity, and $p_{R_a}$ and $p_{R_b}$ are the probability distributions region ($R_a$) and ($R_b$) respectively. If the intensities of both regions are statistically independent, then GCNR will be equals to one, whereas, if they completely overlap then GCNR will be zero \cite{rodriguez2019generalized}. 

 In addition to image quality, we also compared the reconstruction time of proposed method.

\subsection{Network training}
 In the supervised learning of the network for comparative study, 
 match image pairs are used to minimize the $l_1$ loss and SSIM loss between target and the network output.
 For unsupervised learning, the loss function defined in \eqref{eq:OTcycleGAN} is minimized.

 Both the supervised and unsupervised methods were implemented using Python on TensorFlow platform \cite{abadi2016tensorflow}. For parameter optimization the Adam optimizer is used \cite{kingma2014adam}. For supervised training the default values of adam were used, while for unsupervised case the learning rate is linearly changed from $5\times10^{-4}$ to $1\times10^{-4}$ in $200$ epochs.

\begin{figure}[!hbt]
	\centerline{\includegraphics*[width=9cm]{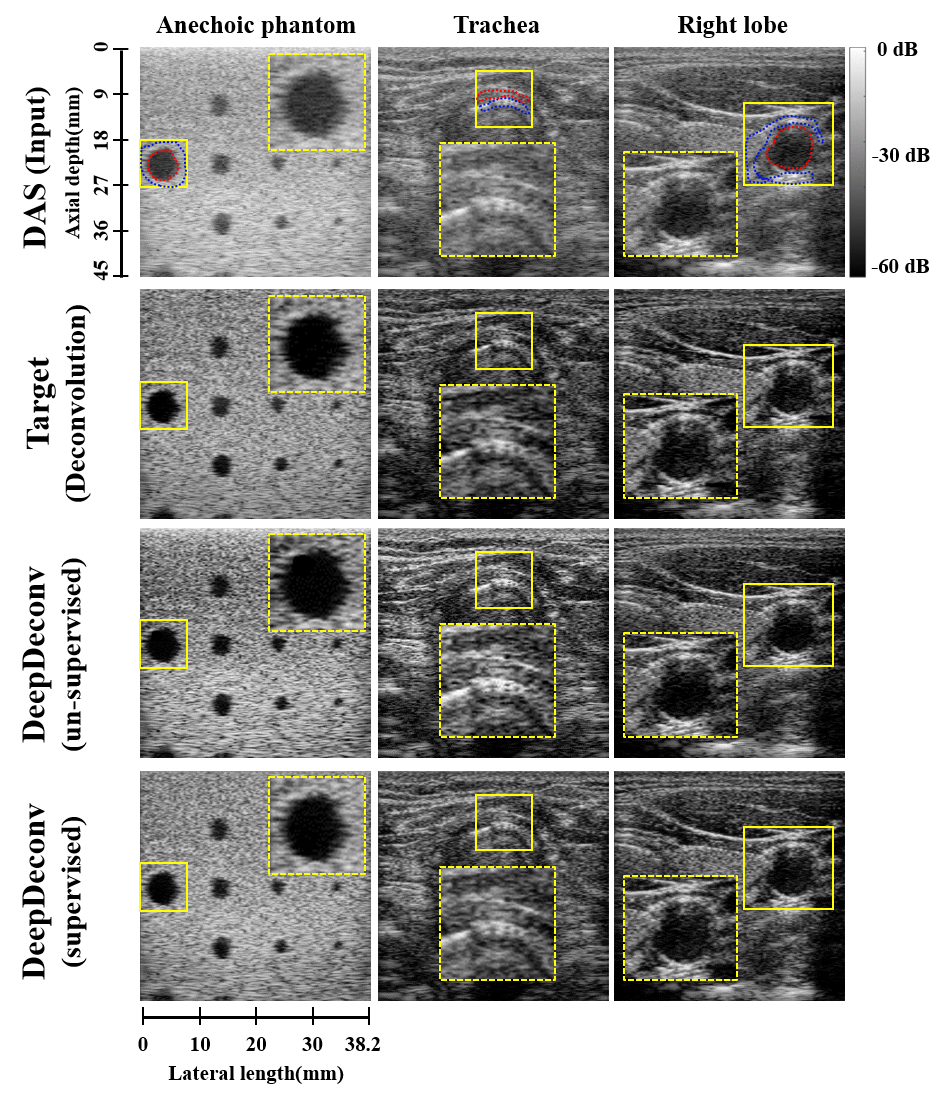}}
	\caption{Deconvolution results on fully-sampled RF data. B-Mode images from tissue mimicking phantom (left), and from \textit{in-vivo} data of carotid region (center \& right).}
	\label{fig:results_deconv_full}	
\end{figure} 
\begin{figure*}[!hbt]
	\centerline{\includegraphics*[width=16cm]{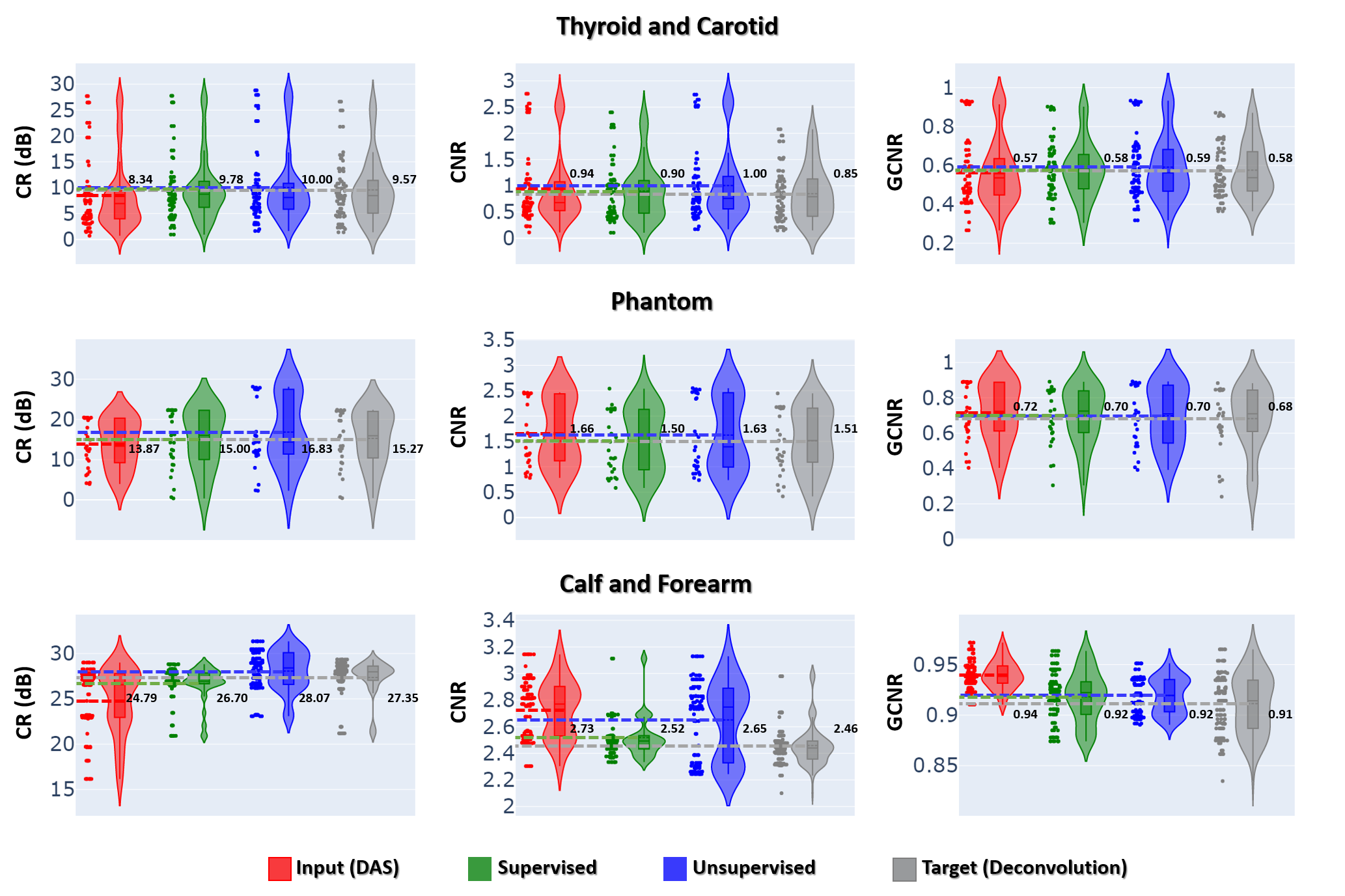}}
	\caption{Comparison of performance metrics by various methods.}
	\label{fig:dist_deepDeconv}	
\end{figure*} 

\begin{figure}[!hbt]
	\centerline{\includegraphics*[width=9cm]{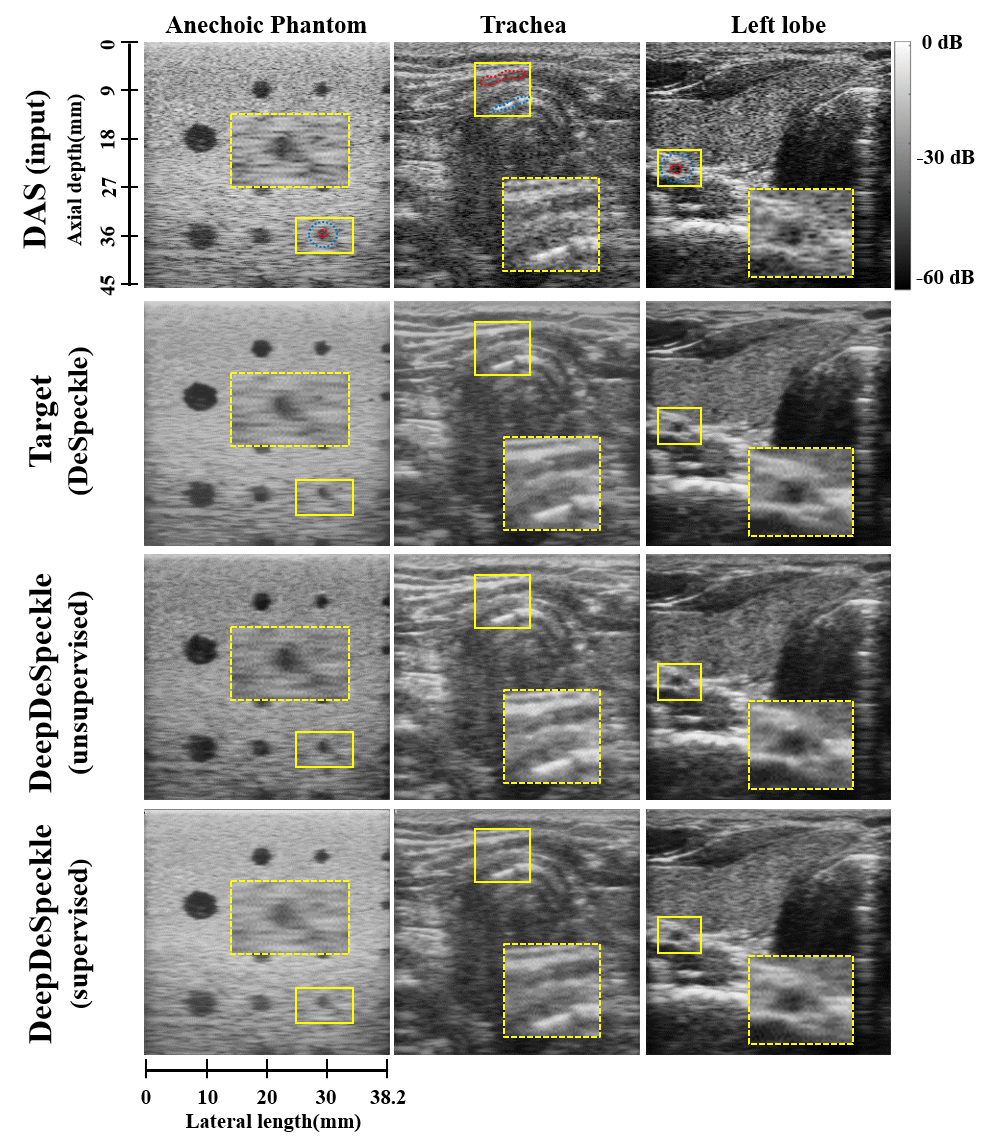}}
	\caption{Speckle removal results on fully-sampled RF data. B-Mode images from tissue mimicking phantom (left), and from \textit{in-vivo} data of carotid region (center \& right).}
	\label{fig:results_despeckle_full}	
\end{figure} 
\begin{figure*}[!hbt]
	\centerline{\includegraphics*[width=16cm]{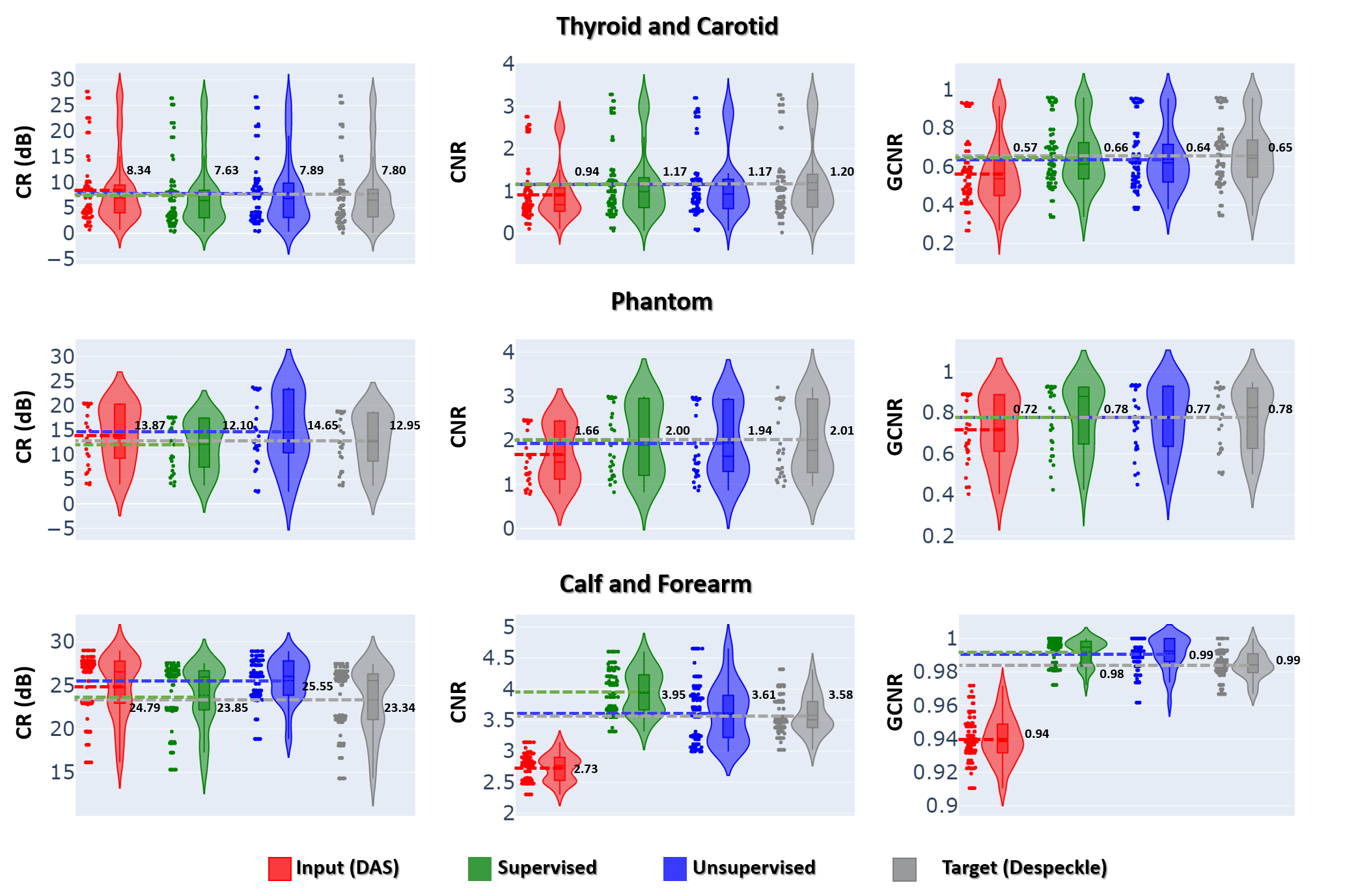}}
	\caption{Comparison of performance metrics by various methods.}
	\label{fig:dist_das_despeckle}	
\end{figure*}

\section{Results }
\label{sec:results}

In the following, we verify the performance of the algorithm for the following experiments:
\begin{enumerate}
\item Deconvolution ultrasound.
\item Speckle noise removal.
\item Planewave image enhancement using high quality focused b-mode target.
\item MLA block artifact and noise removal in cardiac imaging.
\item High quality ultrasound imaging from sub-sampled channel data.
\end{enumerate}
 For the calculation of contrast metrics two regions are selected as highlighted with red and blue dotted lines in respective figures. The same regions are magnified as inset figure for better visualization.

\subsubsection{Deconvolution}

Fig.~\ref{fig:results_deconv_full} show example results from \textit{in-vivo} and phantom scans. From the figures it can be easily seen that the deconvoluted target images have better contrast and anatomical structures are quite prominent compared to the input DAS images.  
Furthermore, both supervised and unsupervised deep learning method successfully learn the deconvolution filtering and improve the visual quality of das images.

For quantitative comparison, the improvement in visual quality is quantified and it is prominent in the contrast distribution plots shown in Fig.~\ref{fig:dist_deepDeconv}. In particular, using unsupervised learning method,  CR values are comparable to the supervised method. As expected, contrary to CR, the CNR and GCNR values are reduced in deconvolution targets, this is because the deconvolution enhances resolution at the cost of noisy high-frequency components. Therefore, the contrast to noise ratio drops substantially, however it is worth noting that at the same CNR and GCNR the CR gain in unsupervised method is much higher compared to label and supervised method. In particular, compared to input (DAS) the proposed unsupervised method recover $3.28$ dB, $2.96$ dB, and $1.66$ dB better CR in Calf and Forearm, Phantom and Thyroid and Carotid regions scans respectively, which is $71.60\%$, $162.02\%$, and $16.11\%$ higher than supervised method.

\subsubsection{Speckle removal}
In this experiment, we perform the speckle de-noising from DAS images using NLLR \cite{NLLR_Despeckle}, supervised, and
unsupervised learning methods.  In Fig.~\ref{fig:results_despeckle_full} one phantom and two in-vivo examples are shown. As for comparison, DAS input images are filtered using NNLR\cite{NLLR_Despeckle} method. Compared to the DAS images, the speckle noise in output images is noticeably reduced. The granular patterns in output images are well suppressed, and resultant images from both supervised and unsupervised
learning methods are similar to the target speckle free images. Here it is noteworthy to point-out that the reconstruction time of deep
learning methods is several magnitude lower than the NLLR\cite{NLLR_Despeckle} method, and unlike NLLR no parameter tuning is needed.

In order to quantify the performance gain, we utilized the same performance metrics used in deconvolution experiments, and the results are shown in Fig.~\ref{fig:dist_das_despeckle}. Interestingly,  the CNR values in despeckle methods are significantly improved.
The reason for high CNR and GCNR is that the  despeckle methods suppresses the noises while maintain the contrast and structural details.

 In particular, compared to input DAS image, the proposed unsupervised method enhance the CNR by $0.89$ units, $0.28$ units, and $0.23$ units in Calf and Forearm, Phantom and Thyroid and Carotid regions scans respectively, which is comparable to supervised methods which shows $1.22$, $0.34$, and $0.23$ units gain, and NLLR \cite{NLLR_Despeckle} method which show $0.85$, $0.35$, and $0.26$ units gain in Calf and Forearm, Phantom and Thyroid and Carotid regions scans, respectively.

\subsubsection{Planewave image enhancement using high quality focused B-mode target.}

\begin{figure*}[!hbt]
	\centerline{\includegraphics*[width=\textwidth]{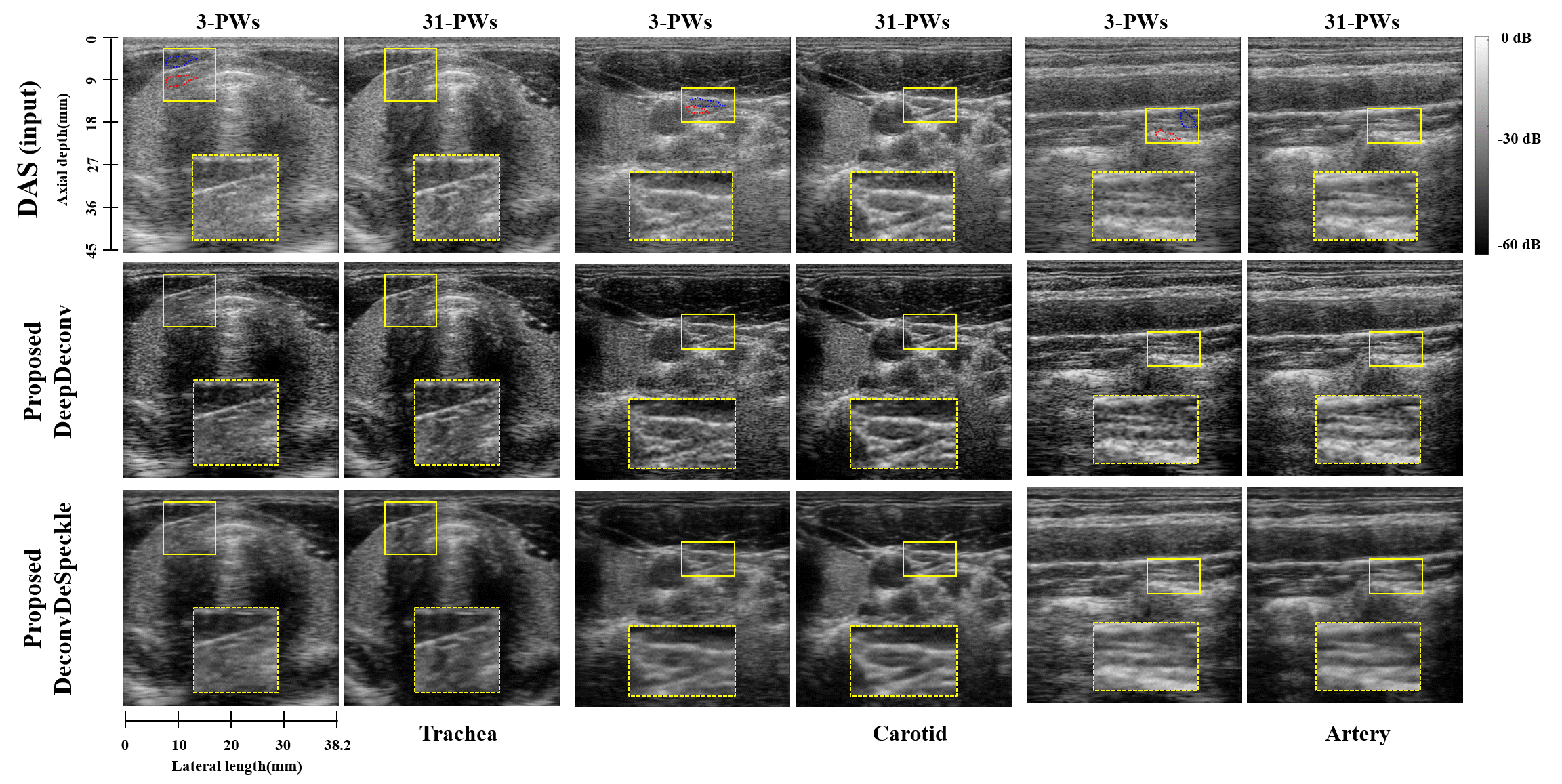}}
	\caption{Planewave ultrasound images enhancement. B-Mode images from \textit{in-vivo} data of carotid regions.}
	\label{fig:DeconvDespeckle_Full_Bmode_PWI}	
\end{figure*} 

In this experiment, we design a model to improve the quality of planewave images using high quality focused B-mode images as target images.  In particular, different sets of PWs were used to generate low quality input images. In  Fig.~\ref{fig:DeconvDespeckle_Full_Bmode_PWI}, three example results are shown. From the figure it can be seen that the contrast of the input images has been improved. 
It is remarkable that  for each type of enhancement, a single trained model is used for all acceleration factors (PWs combinations). In Table~\ref{tbl:results_PWI_enhacement}, three different measures of contrast are shown. From the results it can be clearly seen that the statistics of the output images are substantially improved. When
the deconvoluted B-mode images are used as target distribution,  on average there is a $3.60$ dB gain in CR;
when deconvoluted and speckle removed B-mode images are used as targets,
 on average there is a $4.20$ dB, $0.565$ units and $0.04$ units gain in terms of CR, CNR and GCNR respectively.

\begin{table}[!hbt]
	\centering
	\caption{Comparison of performance statistics on test data in the planewave image enhancement.}
	\label{tbl:results_PWI_enhacement}
	\resizebox{9cm}{!}{
		\begin{tabular}{c|ccc|ccc|ccc}
			\hline
			number of & \multicolumn{3}{c|}{CR (dB)} & \multicolumn{3}{c|}{{CNR}} & \multicolumn{3}{c}{{GCNR}}  \\
			{PWs} & \textit{a} &  \textit{b} &  \textit{c} & \textit{a} &  \textit{b} &  \textit{c} & \textit{a} &  \textit{b} &  \textit{c} \\ \hline\hline
			3 & 13.36 & 17.42 & 17.73 & 2.05 & 2.09 & 2.55 & 0.8405 & 0.8434 & 0.8904 \\
			7 & 15.24 & 18.64 & 19.81 & 2.29 & 2.29 & 2.86 & 0.8739 & 0.8742 & 0.9162 \\
			11 & 16.23 & 19.65 & 20.54 & 2.48 & 2.46 & 3.11 & 0.8962 & 0.8949 & 0.9312 \\
			31 & 17.17 & 20.69 & 20.73 & 2.65 & 2.62 & 3.21 & 0.9164 & 0.9165 & 0.9388 \\  \hline
		\end{tabular}
	}
	\\
	\tiny{$^a$ Input, $^b$ Deconvolution targets, $^c$ Deconvolution + despeckle targets}
	\vspace{0cm}
\end{table}

\subsubsection{High quality accelerated Echocardiograph}
\begin{figure*}[!hbt]
	\centerline{\includegraphics*[width=\textwidth]{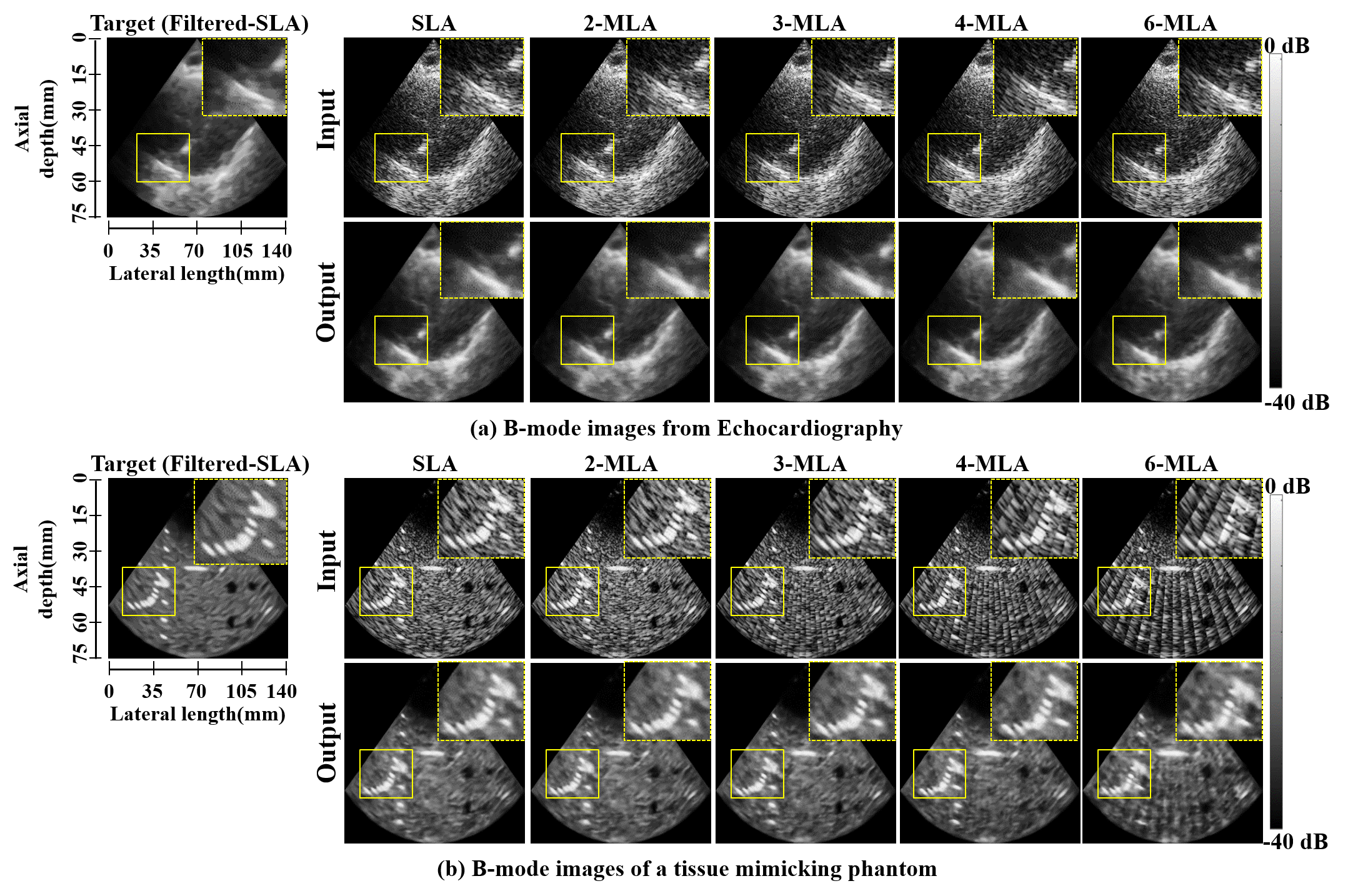}}
	\caption{Reconstruction of high quality accelerated imaging: (a) B-modes images from echocardiograph. (b) B-Mode images from a tissue mimicking phantom.}
	\label{fig:results_cardiac_MLA}	
\end{figure*}

In Fig.~\ref{fig:results_cardiac_MLA}(a) reconstruction results of conventional MLA (referred to as input) and the proposed methods are compared for different number of acceleration factors (transmit events),
which are referred as SLA, 2-MLA, 3-MLA, 4-MLA, and 6-MLA respectively.  For better visualization of the noise suppression effect of our proposed method, a region selected in target image is zoomed out and shown as an inset image. From the example results, it is quite evident that the reconstructed images are very much similar to the target image and have less noise/artifacts compared to input image. Our method sufficiently enhance the visual quality of the input images by eliminating both the speckle and block artifacts for all acceleration factors. It is noteworthy to point out that the proposed method is trained in an unsupervised fashion and a single one-time trained model is used for all acceleration factors.  From reconstructions error statistics in Table~\ref{tbl:results_cardiac_MLA}, it is evident that the quality degradation in input images is much higher than the output images. In particular, on average there is  $0.1372$ gain in GCNR. 

Here we would like to emphasize that the proposed method is based on a single universal model which is one-time trained for multi-tasks i.e., blocking artifact and speckle noise removal and it works for variable MLA schemes without retraining.

 Since in real in-vivo case it is  difficult to decide between anomalies and true structures, to ensure structural preservation we provided additional results using tissue mimicking phantom in Fig.~\ref{fig:results_cardiac_MLA}(b). The results confirm that our method can accurately recover the phantom for most cases. However, with higer acceleration factors e.g., $6$-MLA the block artefacts are becoming prominent and recovery to target quality is not ideal.

Apart from reconstruction quality improvement one major advantage of our method is the fast reconstruction time. 
This is especially important for real time echocardiography that requires fast image reconstruction. 
 Once a model is successfully trained, on average reconstruction time for a single image is around $7.92$ (milliseconds), which is same for all acceleration factors and it could further reduce by optimized implementation.

\begin{table}[!hbt]
	\centering
	\caption{Comparison of performance statistics on test data for noise and block artifact removal from MLA.}
	\label{tbl:results_cardiac_MLA}
	\resizebox{4cm}{!}{
		\begin{tabular}{c|cc}
			\hline
			SLA/MLA & \multicolumn{2}{c}{GCNR} \\ 
			{factor} & \textit{Input} &  \textit{Output} \\ \hline\hline 
			SLA & 0.8221 & 0.9422  \\
			2-MLA & 0.8173 & 0.9380\\
			3-MLA & 0.8072 & 0.9361 \\
			4-MLA & 0.7797 & 0.9313 \\
			6-MLA & 0.7263 & 0.8911 \\
			\hline
		\end{tabular}
	}
	\vspace{0cm}
\end{table}

\subsubsection{High Quality Ultrasound Imaging From Sub-Sampled Channel Data}

In this experiment, six sets of RF data at different down-sampling rates is generated. For each case a separate image is generated, mimicking low-quality/low-powered imaging conditions. For all sub-sampling configurations a single universal model is used, i.e., no separate training is performed for individual sub-sampling case.

\begin{figure}[!hbt]
	\centerline{\includegraphics*[width=9cm]{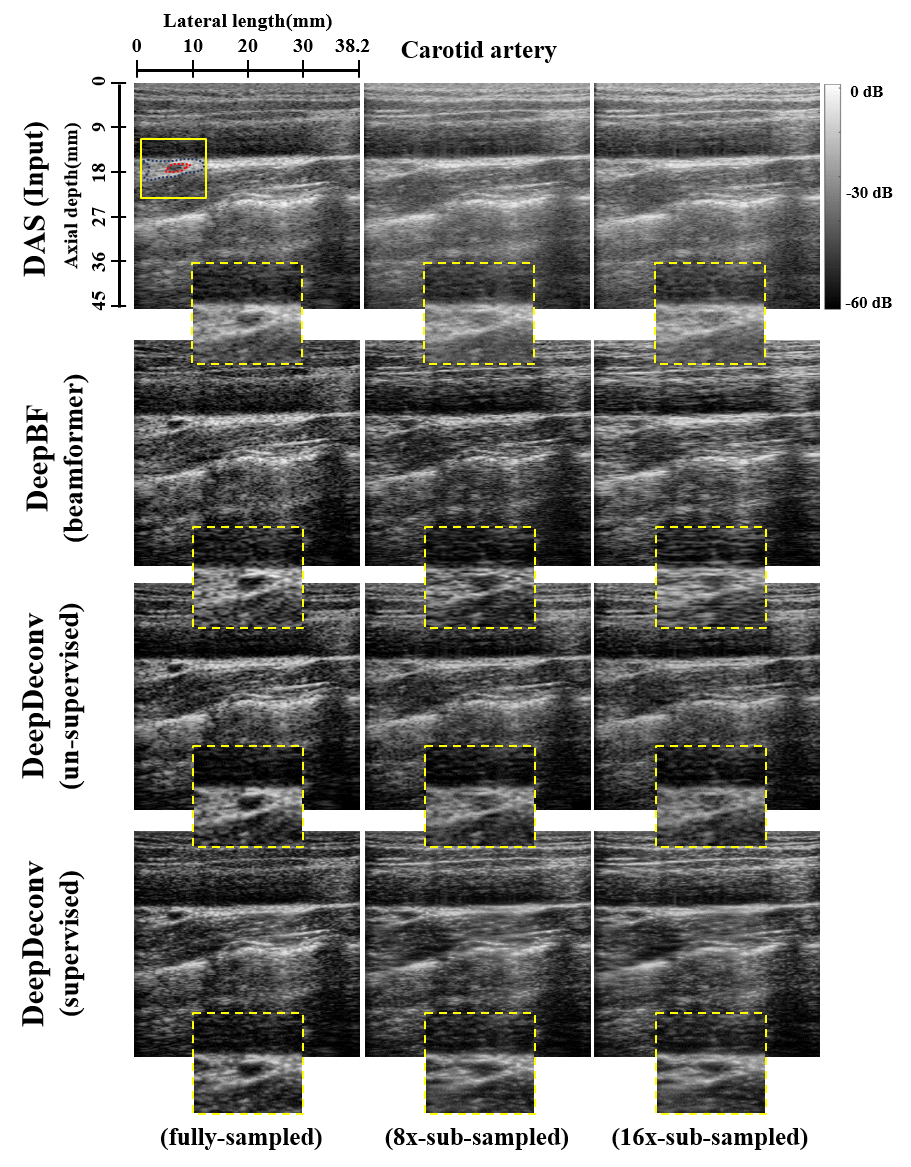}}
	\caption{Reconstruction of high quality ultrasound image from sub-sampled RF data. B-Mode images from \textit{in-vivo} data of carotid artery region.}
	\label{fig:results_deconv_sub}	
\end{figure}

Fig.~\ref{fig:results_deconv_sub} shows the results on example images generated for proposed unsupervised learning method and DAS method using sub-sampled RF data. The trained models are evaluated for standard quality measures.  Overall a single model (one-time trained using either supervised/unsupervised approach) produces significant performance gain in terms of contrast, and resolution for all RF sub-sampling configurations and the performance is comparable with the supervised learning (see Table~\ref{tbl:deconv_sub}).  

\begin{table}[!hbt]
	\centering
	\caption{Comparison of performance statistics on test data for compressive deconvolution ultrasound.}
	\label{tbl:deconv_sub}
	\resizebox{0.4\textwidth}{!}{
		\begin{tabular}{c|ccc}
			\hline
			sub-sampling & \multicolumn{3}{c}{CR (dB)}\\ 
			{factor} & \textit{Input} & \textit{Supervised} & \textit{Unsupervised} 
			\\ \hline\hline
			1 & 8.32 & 9.80 & 11.06 \\ 
			2 & 7.10 & 9.24 & 9.83 \\ 
			4 & 6.39 & 8.54 & 8.64 \\ 
			8 & 6.13 & 8.39 & 7.87 \\
			16 & 6.09 & 8.34 & 7.35\\
			  \hline
		\end{tabular}
	}
	\vspace*{-0.3cm}
\end{table}

\section{Conclusion}
\label{sec:conclusion}
Medical ultrasound imaging is prone to variety of artifacts such as resolution and contrast loss due to insufficient measurement, speckle noise, etc. These artifacts are the major reasons of quality degradation in ultrasound imaging.  
 To address this issue, we proposed  a robust unsupervised deep learning approach that can help generate high-quality US images from low quality noisy images.   Compared to the black-box approaches, our approach was derived based on the rigorous formulation of unsupervised learning using optimal
 transport theory, so with proper training, the method provided reliable reconstruction results without creating any artificial features.
 Since our method does not require paired data for training,  the method can be applied for various US image enhancement applications, providing
 an important platform for further investigation.
 

\end{document}

%% file: packages.tex
\usepackage{outline}
\usepackage{pmgraph}
\usepackage[normalem]{ulem}
\usepackage[utf8]{inputenc}
\usepackage{amssymb}
\usepackage{hyperref}
\usepackage{amsmath}
\usepackage{graphicx}
\usepackage{times}
\usepackage{xcolor}
\usepackage{xspace}
\usepackage[colorinlistoftodos]{todonotes} 
\usepackage{cite}
\usepackage{bm} 
\usepackage{url}

\usepackage{epstopdf}
\AppendGraphicsExtensions{.tiff}
\graphicspath{{fig/}} 

\usepackage{epsfig}
\usepackage{tikz}
\usetikzlibrary{spy}
\usepackage{algpseudocode}
\usepackage{algorithm}
\usepackage{mathrsfs}



%% file: macros.tex
\def\QED{~\rule[-1pt]{5pt}{5pt}\par\medskip}

\long\def\comment#1{} 

\newcommand{\Xc}{\mathcal{X}}
\newcommand{\Yc}{\mathcal{Y}}

\newcommand{\Rd}{{\mathbb R}}

\newcommand{\Ed}{{{\mathbb E}}}

\newcommand{\beq}{\begin{equation}}
\newcommand{\eeq}{\end{equation}}
\newcommand{\beqa}{\begin{eqnarray}}
\newcommand{\eeqa}{\end{eqnarray}}

%% file: main_final_submission_TUFFC1.bbl
\begin{thebibliography}{10}
	\providecommand{\url}[1]{#1}
	\csname url@samestyle\endcsname
	\providecommand{\newblock}{\relax}
	\providecommand{\bibinfo}[2]{#2}
	\providecommand{\BIBentrySTDinterwordspacing}{\spaceskip=0pt\relax}
	\providecommand{\BIBentryALTinterwordstretchfactor}{4}
	\providecommand{\BIBentryALTinterwordspacing}{\spaceskip=\fontdimen2\font plus
		\BIBentryALTinterwordstretchfactor\fontdimen3\font minus
		\fontdimen4\font\relax}
	\providecommand{\BIBforeignlanguage}[2]{{%
			\expandafter\ifx\csname l@#1\endcsname\relax
			\typeout{** WARNING: IEEEtran.bst: No hyphenation pattern has been}%
			\typeout{** loaded for the language `#1'. Using the pattern for}%
			\typeout{** the default language instead.}%
			\else
			\language=\csname l@#1\endcsname
			\fi
			#2}}
	\providecommand{\BIBdecl}{\relax}
	\BIBdecl
	
	\bibitem{TimeReversal1}
	M.~Fink, ``Time reversal of ultrasonic fields. i. basic principles,''
	\emph{IEEE Transactions on Ultrasonics, Ferroelectrics, and Frequency
		Control}, vol.~39, no.~5, pp. 555--566, Sep. 1992.
	
	\bibitem{montaldo2009coherent}
	G.~Montaldo, M.~Tanter, J.~Bercoff, N.~Benech, and M.~Fink, ``Coherent
	plane-wave compounding for very high frame rate ultrasonography and transient
	elastography,'' \emph{IEEE transactions on ultrasonics, ferroelectrics, and
		frequency control}, vol.~56, no.~3, pp. 489--506, 2009.
	
	\bibitem{zhang2020despeckle}
	J.~Zhang and Y.~Cheng, ``Despeckle filters for medical ultrasound images,'' in
	\emph{Despeckling Methods for Medical Ultrasound Images}.\hskip 1em plus
	0.5em minus 0.4em\relax Springer, 2020, pp. 19--45.
	
	\bibitem{NLLR_Despeckle}
	L.~{Zhu}, C.~{Fu}, M.~S. {Brown}, and P.~{Heng}, ``A non-local low-rank
	framework for ultrasound speckle reduction,'' in \emph{2017 IEEE Conference
		on Computer Vision and Pattern Recognition (CVPR)}, July 2017, pp. 493--501.
	
	\bibitem{coupe2009nonlocal}
	P.~Coup{\'e}, P.~Hellier, C.~Kervrann, and C.~Barillot, ``Nonlocal means-based
	speckle filtering for ultrasound images,'' \emph{IEEE transactions on image
		processing}, vol.~18, no.~10, pp. 2221--2229, 2009.
	
	\bibitem{chen2015compressive}
	Z.~Chen, A.~Basarab, and D.~Kouam{\'e}, ``Compressive deconvolution in medical
	ultrasound imaging,'' \emph{IEEE transactions on medical imaging}, vol.~35,
	no.~3, pp. 728--737, 2015.
	
	\bibitem{jensen1992deconvolution}
	J.~A. Jensen, ``Deconvolution of ultrasound images,'' \emph{Ultrasonic
		imaging}, vol.~14, no.~1, pp. 1--15, 1992.
	
	\bibitem{7565583}
	J.~{Duan}, H.~{Zhong}, B.~{Jing}, S.~{Zhang}, and M.~{Wan}, ``Increasing axial
	resolution of ultrasonic imaging with a joint sparse representation model,''
	\emph{IEEE Transactions on Ultrasonics, Ferroelectrics, and Frequency
		Control}, vol.~63, no.~12, pp. 2045--2056, Dec 2016.
	
	\bibitem{schretter2017ultrasound}
	C.~Schretter, S.~Bundervoet, D.~Blinder, A.~Dooms, J.~{D’hooge}, and
	P.~Schelkens, ``Ultrasound imaging from sparse rf samples using system point
	spread functions,'' \emph{IEEE transactions on ultrasonics, ferroelectrics,
		and frequency control}, vol.~65, no.~3, pp. 316--326, 2017.
	
	\bibitem{kang2017deep}
	E.~Kang, J.~Min, and J.~C. Ye, ``A deep convolutional neural network using
	directional wavelets for low-dose x-ray ct reconstruction,'' \emph{Medical
		Physics}, vol.~44, no.~10, 2017.
	
	\bibitem{ye2017deep}
	J.~C. Ye, Y.~Han, and E.~Cha, ``Deep convolutional framelets: A general deep
	learning framework for inverse problems,'' \emph{SIAM Journal on Imaging
		Sciences}, vol.~11, no.~2, pp. 991--1048, 2018.
	
	\bibitem{han2017deep}
	Y.~S. Han, J.~Yoo, and J.~C. Ye, ``Deep learning with domain adaptation for
	accelerated projection reconstruction mr,'' \emph{arXiv preprint
		arXiv:1703.01135}, 2017.
	
	\bibitem{aggarwal2017modl}
	H.~K. Aggarwal, M.~P. Mani, and M.~Jacob, ``{MoDL}: Model based deep learning
	architecture for inverse problems,'' \emph{arXiv preprint arXiv:1712.02862},
	2017.
	
	\bibitem{khan2019deep}
	S.~Khan, J.~Huh, and J.~C. Ye, ``Deep learning-based universal beamformer for
	ultrasound imaging,'' in \emph{International Conference on Medical Image
		Computing and Computer-Assisted Intervention}.\hskip 1em plus 0.5em minus
	0.4em\relax Springer, Cham, 2019, pp. 619--627.
	
	\bibitem{wurfl2016deep}
	T.~W{\"u}rfl, F.~C. Ghesu, V.~Christlein, and A.~Maier, ``Deep learning
	computed tomography,'' in \emph{International Conference on Medical Image
		Computing and Computer-Assisted Intervention}.\hskip 1em plus 0.5em minus
	0.4em\relax Springer, 2016, pp. 432--440.
	
	\bibitem{yoon2018efficient}
	Y.~H. Yoon, S.~Khan, J.~Huh, J.~C. Ye \emph{et~al.}, ``Efficient b-mode
	ultrasound image reconstruction from sub-sampled rf data using deep
	learning,'' \emph{IEEE transactions on medical imaging}, 2018.
	
	\bibitem{khan2020adaptive}
	S.~Khan, J.~Huh, and J.~C. Ye, ``Adaptive and compressive beamforming using
	deep learning for medical ultrasound,'' \emph{IEEE Transactions on
		Ultrasonics, Ferroelectrics, and Frequency Control}, 2020.
	
	\bibitem{feigin2018deep}
	M.~Feigin, D.~Freedman, and B.~W. Anthony, ``A deep learning framework for
	single sided sound speed inversion in medical ultrasound,'' \emph{arXiv
		preprint arXiv:1810.00322}, 2018.
	
	\bibitem{nair2018fully}
	A.~A. Nair, M.~R. Gubbi, T.~D. Tran, A.~Reiter, and M.~A.~L. Bell, ``A fully
	convolutional neural network for beamforming ultrasound images,'' in
	\emph{2018 IEEE International Ultrasonics Symposium (IUS)}.\hskip 1em plus
	0.5em minus 0.4em\relax IEEE, 2018, pp. 1--4.
	
	\bibitem{khan2019universal}
	S.~Khan, J.~Huh, and J.~C. Ye, ``Universal plane-wave compounding for high
	quality us imaging using deep learning,'' in \emph{2019 IEEE International
		Ultrasonics Symposium (IUS)}.\hskip 1em plus 0.5em minus 0.4em\relax IEEE,
	2019, pp. 2345--2347.
	
	\bibitem{jafari2020cardiac}
	M.~H. Jafari, H.~Girgis, N.~Van~Woudenberg, N.~Moulson, C.~Luong, A.~Fung,
	S.~Balthazaar, J.~Jue, M.~Tsang, P.~Nair \emph{et~al.}, ``Cardiac
	point-of-care to cart-based ultrasound translation using constrained
	cyclegan,'' \emph{International Journal of Computer Assisted Radiology and
		Surgery}, pp. 1--10, 2020.
	
	\bibitem{vedula2018high}
	S.~Vedula, O.~Senouf, G.~Zurakhov, A.~Bronstein, M.~Zibulevsky,
	O.~Michailovich, D.~Adam, and D.~Gaitini, ``High quality ultrasonic
	multi-line transmission through deep learning,'' in \emph{International
		Workshop on Machine Learning for Medical Image Reconstruction}.\hskip 1em
	plus 0.5em minus 0.4em\relax Springer, 2018, pp. 147--155.
	
	\bibitem{nair2019generative}
	A.~A. Nair, T.~D. Tran, A.~Reiter, and M.~A.~L. Bell, ``A generative
	adversarial neural network for beamforming ultrasound images: Invited
	presentation,'' in \emph{2019 53rd Annual Conference on Information Sciences
		and Systems (CISS)}.\hskip 1em plus 0.5em minus 0.4em\relax IEEE, 2019, pp.
	1--6.
	
	\bibitem{yu2018deep}
	Z.~Yu, E.-L. Tan, D.~Ni, J.~Qin, S.~Chen, S.~Li, B.~Lei, and T.~Wang, ``A deep
	convolutional neural network-based framework for automatic fetal facial
	standard plane recognition,'' \emph{IEEE journal of biomedical and health
		informatics}, vol.~22, no.~3, pp. 874--885, 2018.
	
	\bibitem{ozkan2018inverse}
	E.~Ozkan, V.~Vishnevsky, and O.~Goksel, ``Inverse problem of ultrasound
	beamforming with sparsity constraints and regularization,'' \emph{IEEE
		transactions on ultrasonics, ferroelectrics, and frequency control}, vol.~65,
	no.~3, pp. 356--365, 2018.
	
	\bibitem{cohen2018sparse}
	R.~Cohen and Y.~C. Eldar, ``Sparse convolutional beamforming for ultrasound
	imaging,'' \emph{IEEE transactions on ultrasonics, ferroelectrics, and
		frequency control}, vol.~65, no.~12, pp. 2390--2406, 2018.
	
	\bibitem{huang2019mimicknet}
	O.~Huang, W.~Long, N.~Bottenus, G.~E. Trahey, S.~Farsiu, and M.~L. Palmeri,
	``Mimicknet, matching clinical post-processing under realistic black-box
	constraints,'' in \emph{2019 IEEE International Ultrasonics Symposium
		(IUS)}.\hskip 1em plus 0.5em minus 0.4em\relax IEEE, 2019, pp. 1145--1151.
	
	\bibitem{Reference:sim2019optimal}
	B.~Sim, G.~Oh, S.~Lim, and J.~C. Ye, ``Optimal transport, cycle{GAN}, and
	penalized {LS} for unsupervised learning in inverse problems,'' \emph{arXiv
		preprint arXiv:1909.12116}, 2019.
	
	\bibitem{villani2008optimal}
	C.~Villani, \emph{Optimal transport: old and new}.\hskip 1em plus 0.5em minus
	0.4em\relax Springer Science \& Business Media, 2008, vol. 338.
	
	\bibitem{peyre2019computational}
	G.~Peyr{\'e}, M.~Cuturi \emph{et~al.}, ``Computational optimal transport,''
	\emph{Foundations and Trends in Machine Learning}, vol.~11, no. 5-6, pp.
	355--607, 2019.
	
	\bibitem{kullback1997information}
	S.~Kullback, \emph{Information theory and statistics}.\hskip 1em plus 0.5em
	minus 0.4em\relax Courier Corporation, 1997.
	
	\bibitem{arjovsky2017wasserstein}
	M.~Arjovsky, S.~Chintala, and L.~Bottou, ``Wasserstein {GAN},'' \emph{arXiv
		preprint arXiv:1701.07875}, 2017.
	
	\bibitem{zhu2017unpaired}
	J.-Y. Zhu, T.~Park, P.~Isola, and A.~A. Efros, ``Unpaired image-to-image
	translation using cycle-consistent adversarial networks,'' in
	\emph{Proceedings of the IEEE international conference on computer vision},
	2017, pp. 2223--2232.
	
	\bibitem{miyato2018spectral}
	T.~Miyato, T.~Kataoka, M.~Koyama, and Y.~Yoshida, ``Spectral normalization for
	generative adversarial networks,'' \emph{arXiv preprint arXiv:1802.05957},
	2018.
	
	\bibitem{gulrajani2017improved}
	I.~Gulrajani, F.~Ahmed, M.~Arjovsky, V.~Dumoulin, and A.~C. Courville,
	``Improved training of {W}asserstein {GAN}s,'' in \emph{Advances in neural
		information processing systems}, 2017, pp. 5767--5777.
	
	\bibitem{lim2019cyclegan}
	S.~Lim, S.-E. Lee, S.~Chang, and J.~C. Ye, ``Cycle{GAN} with a blur kernel for
	deconvolution microscopy: Optimal transport geometry,'' \emph{IEEE Trans. on
		Computational Imaging (in press), also available as arXiv preprint
		arXiv:1908.09414}, 2020.
	
	\bibitem{mao2017least}
	X.~Mao, Q.~Li, H.~Xie, R.~Y. Lau, Z.~Wang, and S.~Paul~Smolley, ``Least squares
	generative adversarial networks,'' in \emph{Proceedings of the IEEE
		International Conference on Computer Vision}, 2017, pp. 2794--2802.
	
	\bibitem{taxt1999noise}
	T.~Taxt and G.~V. Frolova, ``Noise robust one-dimensional blind deconvolution
	of medical ultrasound images,'' \emph{IEEE transactions on ultrasonics,
		ferroelectrics, and frequency control}, vol.~46, no.~2, pp. 291--299, 1999.
	
	\bibitem{yu2012blind}
	C.~Yu, C.~Zhang, and L.~Xie, ``A blind deconvolution approach to ultrasound
	imaging,'' \emph{IEEE transactions on ultrasonics, ferroelectrics, and
		frequency control}, vol.~59, no.~2, pp. 271--280, 2012.
	
	\bibitem{jensen1991estimation}
	J.~A. Jensen, ``Estimation of pulses in ultrasound b-scan images,'' \emph{IEEE
		transactions on medical imaging}, vol.~10, no.~2, pp. 164--172, 1991.
	
	\bibitem{jirik2008two}
	R.~Jirik and T.~Taxt, ``Two-dimensional blind bayesian deconvolution of medical
	ultrasound images,'' \emph{IEEE transactions on ultrasonics, ferroelectrics,
		and frequency control}, vol.~55, no.~10, pp. 2140--2153, 2008.
	
	\bibitem{zhang2012matrix}
	D.~Zhang, Y.~Hu, J.~Ye, X.~Li, and X.~He, ``Matrix completion by truncated
	nuclear norm regularization,'' in \emph{2012 IEEE Conference on Computer
		Vision and Pattern Recognition}.\hskip 1em plus 0.5em minus 0.4em\relax IEEE,
	2012, pp. 2192--2199.
	
	\bibitem{gu2014weighted}
	S.~Gu, L.~Zhang, W.~Zuo, and X.~Feng, ``Weighted nuclear norm minimization with
	application to image denoising,'' in \emph{Proceedings of the IEEE conference
		on computer vision and pattern recognition}, 2014, pp. 2862--2869.
	
	\bibitem{tanter2014ultrafast}
	M.~Tanter and M.~Fink, ``Ultrafast imaging in biomedical ultrasound,''
	\emph{IEEE transactions on ultrasonics, ferroelectrics, and frequency
		control}, vol.~61, no.~1, pp. 102--119, 2014.
	
	\bibitem{gasse2017high}
	M.~Gasse, F.~Millioz, E.~Roux, D.~Garcia, H.~Liebgott, and D.~Friboulet,
	``High-quality plane wave compounding using convolutional neural networks,''
	\emph{IEEE transactions on ultrasonics, ferroelectrics, and frequency
		control}, vol.~64, no.~10, pp. 1637--1639, 2017.
	
	\bibitem{matrone2017high}
	G.~Matrone, A.~Ramalli, A.~S. Savoia, P.~Tortoli, and G.~Magenes, ``High
	frame-rate, high resolution ultrasound imaging with multi-line transmission
	and filtered-delay multiply and sum beamforming,'' \emph{IEEE transactions on
		medical imaging}, vol.~36, no.~2, pp. 478--486, 2017.
	
	\bibitem{wagner2012compressed}
	N.~Wagner, Y.~C. Eldar, and Z.~Friedman, ``Compressed beamforming in ultrasound
	imaging,'' \emph{IEEE Transactions on Signal Processing}, vol.~60, no.~9, pp.
	4643--4657, 2012.
	
	\bibitem{burshtein2016sub}
	A.~Burshtein, M.~Birk, T.~Chernyakova, A.~Eilam, A.~Kempinski, and Y.~C. Eldar,
	``Sub-nyquist sampling and fourier domain beamforming in volumetric
	ultrasound imaging,'' \emph{IEEE Trans. Ultrason., Ferroelectr., Freq.
		Control}, vol.~63, no.~5, pp. 703--716, 2016.
	
	\bibitem{rodriguez2019generalized}
	A.~{Rodriguez-Molares}, O.~M.~H. {Rindal}, J.~{D’hooge}, S.~M{\aa}s{\o}y,
	A.~{Austeng}, M.~A. {Lediju Bell}, and H.~{Torp}, ``The generalized
	contrast-to-noise ratio: A formal definition for lesion detectability,''
	\emph{IEEE Transactions on Ultrasonics, Ferroelectrics, and Frequency
		Control}, vol.~67, no.~4, pp. 745--759, 2020.
	
	\bibitem{abadi2016tensorflow}
	M.~Abadi, P.~Barham, J.~Chen, Z.~Chen, A.~Davis, J.~Dean, M.~Devin,
	S.~Ghemawat, G.~Irving, M.~Isard \emph{et~al.}, ``Tensorflow: A system for
	large-scale machine learning.'' in \emph{OSDI}, vol.~16, 2016, pp. 265--283.
	
	\bibitem{kingma2014adam}
	D.~Kingma and J.~Ba, ``{Adam: A method for stochastic optimization},''
	\emph{arXiv preprint arXiv:1412.6980}, 2014.
	
\end{thebibliography}
